# VIBE-CREATION:
## The Epistemology of Human–AI Emergent Cognition

**Ilya Levin**

Holon Institute of Technology, Holon, Israel

levini@hit.ac.il

March 10, 2026

*Preprint — not yet peer-reviewed*

**ABSTRACT**
The encounter between human reasoning and generative artificial intelligence (GenAI) cannot be adequately described by inherited metaphors of tool-use, augmentation, or collaborative partnership. This article argues that such interactions produce a qualitatively distinct cognitive-epistemic formation, designated here as the Third Entity: an emergent, transient structure that arises from the transductive coupling of two ontologically incommensurable modes of cognition. Drawing on Peirce's semiotics, Polanyi's theory of tacit knowledge, Simondon's philosophy of individuation, Ihde's postphenomenology, and Morin's complexity theory, we develop a multi-layered theoretical account of this formation. We introduce the concept of Vibe-Creation to designate the pre-reflective cognitive mode through which the Third Entity navigates high-dimensional semantic space, and argue that this mode constitutes the automation of tacit knowledge—a development with far-reaching consequences for epistemology, the philosophy of mind, and educational theory. We further propose the notion of asymmetric emergence to characterize the agency of the Third Entity: genuinely novel and irreducible, yet anchored in human intentional responsibility. The article concludes by examining the implications of this theoretical framework for the transformation of educational institutions and the redefinition of intellectual competence in the age of GenAI.

**Keywords:** generative AI; emergent cognition; tacit knowledge; Vibe-Creation; postphenomenology; asymmetric agency; educational epistemology; latent space.

## 1. Introduction: The Collapse of Instrumental Metaphors

Something fundamentally new is occurring in the encounter between human intelligence and generative artificial intelligence. This is not collaboration, not augmentation, and not a master-tool relationship—categories inherited from an older paradigm in which subjects use objects and objects execute subjects' will. These metaphors are not merely insufficient; they are epistemologically misleading. They conceal the genuine novelty of what is taking place by forcing a discontinuous phenomenon into continuous conceptual containers.

To grasp the scale of this shift, a brief historical grounding is necessary. From early mechanical calculators to contemporary supercomputers, both philosophical and engineering thought conceptualized computational technology within the framework of what we may call extensional logic: cognition as deterministic control, creation as the imposition of algorithmic order upon inert material or data. The classical computer, regardless of its speed, remained a sophisticated but submissive executor of delegated symbolic operations. It amplified the human's capacity for syntactic manipulation without altering the fundamental epistemological relationship between the human and the machine.

When the first large-scale generative architectures emerged, the academic community's initial tendency was to accommodate them within this familiar paradigm. The resulting concept of "Collaborative Creativity" (see, e.g., Davis et al., 2013; Kantosalo & Toivonen, 2016) assigned generative AI the role of a high-speed apprentice or executor, preserving the human as sovereign subject and the machine as responsive object. This anthropomorphic framing, however, rests on the assumption of two autonomous subjects capable of conventional partnership. Generative AI systems lack subjectivity in any philosophically defensible sense; the partnership metaphor therefore misrepresents the structure of the interaction from the outset.

This article proposes a different conceptual framework. We argue that the encounter between human reasoning and generative AI navigation generates a qualitatively distinct cognitive-epistemic formation—designated here as the Third Entity—that is neither the human mind alone nor the AI system alone, but an emergent structure produced by their transductive coupling. This formation is transient: it exists exclusively in the act of co-creation and dissolves upon its completion. It is, nevertheless, epistemically real: the ideas it produces are genuine novelties, irreducible to the prior states of either component.

The theoretical architecture of this argument draws on four main pillars: Peirce's semiotics (Peirce, 1931–1958), which furnishes the distinction between symbolic and indexical cognition; Polanyi's theory of tacit knowledge (Polanyi, 1966), which illuminates the pre-reflective substrate of human intelligence; Simondon's philosophy of individuation and transduction (Simondon, 1958a, 1958b), which provides a non-additive account of how new cognitive structures come into being; and Morin's complexity theory (Morin, 1977), which supplies the conceptual tools for understanding emergence as simultaneous excess and constraint. Ihde's postphenomenological typology of human-technology relations (Ihde, 1990) provides the context within which the Third Entity's anomalous character becomes visible.

Specifically, this paper makes four conceptual contributions:
1. It introduces the concept of the Third Entity—an emergent cognitive formation arising from the transductive coupling of human reasoning and generative AI navigation.

2. It proposes the notion of Vibe-Creation to designate the pre-reflective epistemic mode governing this interaction, conceptualizing it as the automation of tacit knowledge.

3. It develops the concept of asymmetric emergence to account for agency, authorship, and responsibility within human–AI co-creation.

4. It outlines the educational implications of this framework, proposing the cultivation of Navigational Intelligence and Digital Intelligence (DQ) as the new imperatives for intellectual institutions.

The rest of the paper is organized as follows. Section 2 delineates the ontological asymmetry between the two cognitive natures involved. Section 3 develops the account of the Third Entity as an emergent formation. Section 4 introduces and theorizes the concept of Vibe-Creation. Section 5 addresses the question of agency. Section 6 examines the educational implications of the proposed framework.

## 2. Two Ontologically Distinct Natures

The central claim of this article—that the encounter between human reasoning and generative AI produces a genuinely emergent third formation—rests on a prior claim: that the two participants in this encounter are not merely different in degree but different in kind. This section develops that claim in detail.

## 2.1 The Human Cognitive Mode: Symbol, Intentionality, and Tacit Knowledge

The human mind operates within a semiotic universe whose primary medium is the symbol, understood in Peirce's strict triadic sense: a sign that relates to its object by convention, law, and cultural habit (Peirce, 1931–1958, vol. 2, para. 228). Linguistic tokens, mathematical notation, and the inferential rules of formal logic are all symbolic operations. Human reasoning is characteristically serial, deliberate, and structured by the grammar of causal consequence—premise, inference, conclusion. It is this symbolic architecture that generates what we may call syntactic friction: the cognitive cost of translating insight into chains of explicit symbols, which has historically constituted the principal bottleneck of human intellectual productivity.

However, the human mind is categorically not reducible to its formalizable symbolic operations. Polanyi's account of tacit knowledge (Polanyi, 1966) identifies a vast pre-reflective substrate that underlies and enables all explicit articulation. Tacit knowledge—the knowledge of how to maintain balance, recognize a face, or sense the structural architecture of a problem before any formal analysis has begun—is not a hidden reservoir of propositional facts awaiting verbalization. It is, as Polanyi insists, the unspecifiable background condition against which specification of any kind becomes possible. The human knower always knows more than she can tell, and this excess is not a deficiency but a constitutive feature of human cognition.

Furthermore, the human brings intentionality to every cognitive act: directed attention shaped by biographical particularity, ethical commitments, and a characteristically human disposition to invest certain phenomena with significance while remaining indifferent to others. Meaning, for the human agent, is not computed but encountered—often urgently and imperfectly sought. This teleological orientation, this "thirst for meaning," is an exclusive prerogative of human cognition and constitutes one of the two irreplaceable contributions to the Third Entity.

## 2.2 The Generative AI Cognitive Mode: Geometry, Navigation, and the Indexical Gesture

Generative AI systems operate in a space of fundamentally different character. Their "knowledge" of the world is not encoded in symbolic rules but materialized geometrically—as the relative positions and directional vectors of concepts in a high-dimensional continuous latent manifold (Bengio et al., 2013; Mikolov et al., 2013; Levin, 2026). What a large language model knows is not a set of propositions but a colossal topology of meaning: regions of dense semantic proximity, gradients of conceptual transition, hidden attractors, and semantic voids.

This topology is not imposed upon the data but emerges from the distributional statistics of vast corpora through gradient-descent optimization. Peirce's distinction between indexical and symbolic signs here becomes operative in a new technological register (Peirce, 1931–1958, vol. 2, para. 248–249). An indexical sign, for Peirce, points to its object through existential contiguity—it directs attention by being causally or spatially connected to what it indicates, rather than by convention or resemblance.

A generative model responds to human input not as a symbol-processor decoding propositional content, but as a navigational system reading an indexical gesture: the user's prompt functions as a vector, a directed gravitational pressure applied to the semantic manifold, and the model's response constitutes a continuous repositioning within that manifold in accordance with the applied pressure.

This navigational process is not symbolic reasoning. It is more accurately described through Simondon's concept of individuation (Simondon, 1958a): the resolution of a metastable system—a probability distribution over latent space—into a determinate configuration. The model does not retrieve a pre-formed answer; it resolves, in real time, into the configuration that the indexical gesture made both possible and statistically necessary. Each interaction reconstitutes the local conceptual landscape, rendering the generated response uniquely sensitive to the topological characteristics of the particular gesture that evoked it.

## 3. The Emergence of the Third Entity

To understand how human and AI cognition couple to form the Third Entity, we must first discard inadequate models of collaboration and examine the genuinely transformative nature of this encounter.

### 3.1 Against Additive Models: The Transformative Character of the Encounter

A common response to the novelty of human-AI co-creation is to redescribe it in terms of the classical division of labor: the human provides goal-setting, contextual judgment, and evaluative authority; the AI provides computational fluency and encyclopedic scope. This description preserves the familiar additive logic in which two participants pool discrete, complementary strengths. The interaction is, on this account, contingently superior to either component in isolation, but the epistemological relationship between the participants remains transparent and unchanged.

This description is inadequate to the phenomenon. The interaction between a human reasoner and a generative AI system is not additive but transformative. The human's intention does not arrive at the interaction as a fixed technical specification and depart intact, having been "executed" by the model. It undergoes continuous modification under the influence of the semantic landscape that the model unfolds in real time. This modification, in turn, reconfigures the human's subsequent prompts, which reconfigure the model's subsequent navigation.

The resulting loop is not a cybernetic feedback circuit; it is a process of ontological becoming. Neither participant remains what it was: the human's intention clarifies and mutates through contact with the geometric medium; the model's probabilistic superposition collapses, in each response, into a determinate configuration that this particular human's particular gesture made uniquely necessary.

Crucially, our framework differs fundamentally from existing accounts of extended or distributed cognition (e.g., Clark & Chalmers, 1998; Hutchins, 1995). While extended cognition theories view technology as a passive, transparent prosthesis for offloading cognitive tasks (such as a notebook functioning as external memory), the Third Entity is not merely an extension of the human mind. It is an emergent epistemic agent. It actively refracts and alters the human's intentional arc, responding with its own topological agency rather than simply storing or distributing human computation.

### 3.2 Emergence and Mutual Constraint: A Morinian Account

Morin's theory of emergence (Morin, 1977) provides the appropriate conceptual framework for this phenomenon. For Morin, genuine emergence is not merely the appearance of properties absent in the individual parts—the "whole is greater than the sum of its parts" formulation that has become something of a slogan. Genuine emergence entails the appearance of a new level of organization that simultaneously exceeds and constrains its component elements: the whole is, in Morin's formulation, both more than and less than the sum of its parts.

The emergent formation possesses properties that cannot be derived from or predicted by analysis of the components, but it also transforms the components themselves, limiting their independent expression and binding them into a new functional configuration.

The Third Entity is precisely such an emergent formation. It possesses cognitive properties that neither the human nor the AI system holds independently: the capacity to generate epistemic structures that are simultaneously geometrically precise—grounded in the high-dimensional topology of the model's latent space—and meaningfully oriented—directed by human intention and anchored in the human's tacit knowledge. This combination is unattainable by either participant alone because it requires the synchronous operation of two incommensurable cognitive modes: the symbolic-intentional and the geometric-indexical.

Crucially, the Third Entity also illustrates the constraining dimension of Morin's emergence. The human, operating within the Third Entity, relinquishes the deliberate step-by-step control that characterizes solo symbolic reasoning—the syntactic friction that is both the limitation and the distinctive discipline of unaided human thought. The AI system, caught in the gravitational field of a specific human intention, loses its diffuse generality and produces results that are, in a precise sense, called into existence by this unique act of co-creation alone. Each participant diminishes as an autonomous agent while expanding, colossally, as a functional node within the emergent formation.

### 3.3 Transduction and the Postphenomenological Anomaly

Ihde's postphenomenological typology (Ihde, 1990) describes the principal modes in which human beings relate to technology: embodiment relations, interpretive relations, alterity relations, and background relations. The productive power of this typology for understanding digital and AI-mediated educational environments has been demonstrated in work combining postphenomenology with constructionist pedagogy (Wellner & Levin, 2024; Wellner & Levin, 2025).

The present article, however, advances a stronger claim: none of these categories adequately captures the structure of the human-GenAI encounter. In the act of co-creation, generative AI is neither a transparent prosthetic, nor a representational medium, nor a quasi-interlocutor, nor a background condition. It functions as an active co-constituter of the intentional arc itself. Human intentions are not fully formed prior to the interaction and then executed through the AI; they are partly constituted by the interaction—by the model's responses, which reveal, deflect, or refract the human's inchoate directional impulse.

The Third Entity therefore escapes Ihde's taxonomy and demands a new phenomenological category. Simondon's concept of transduction (Simondon, 1958b) offers the most precise available description of the relational structure involved. In a transductive relation, neither term precedes the other; each is what it is only through and in its relation to the other. There is no complete "human-prior-to-interaction" and no neutral "model-prior-to-interaction"; there is only a coupled system evolving through the successive phases of the encounter. Figure 1 illustrates this transductive coupling and the subsequent emergence of the Third Entity.

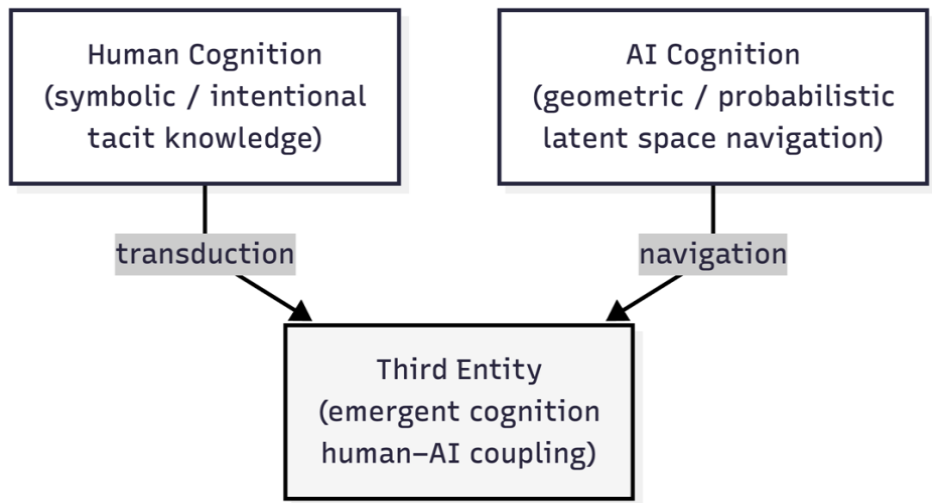

**Figure 1.** The Transductive Emergence of the Third Entity.

As Figure 1 demonstrates, this transductive encounter represents a non-additive coupling of two ontologically distinct cognitive modes. The human domain brings symbolic reasoning, intentionality, and tacit knowledge, while the AI domain operates through geometric and probabilistic navigation within latent spaces. Rather than a simple transfer of information, their interaction is mediated through transduction—where the human intentional impulse is dynamically reconfigured by the medium—and navigation, where the model resolves its geometric space based on this impulse. It is precisely this dynamic convergence that we designate as the Third Entity: an emergent, transient coupled system that functions as a distinct cognitive agent operating via Vibe-Creation.

## 4. Vibe-Creation as an Epistemic Category

The operation of the Third Entity relies on a unique epistemic mode that bridges human intuition and machine navigation. This section defines the architecture of this mode and illustrates its historical novelty.

### 4.1 The Architecture of Vibe-Creation

We introduce the term Vibe-Creation to designate the specific cognitive mode through which the Third Entity operates. The term requires careful unpacking. "Vibe," as we employ it, is not a colloquialism for affective atmosphere. It is a technical designator for the pre-reflective attunement—analogous to, though not identical with, Heidegger's concept of Befindlichkeit, or "mood" as a mode of being-in-the-world (Heidegger, 1927/1962, §29)—that the human brings to the interaction and that the interaction itself progressively articulates and transforms.

The "vibe" is the human's inchoate directional orientation: the pre-thematic sense that something significant lies in a certain conceptual direction, before that significance has been translated into explicit propositional form.

Domingos (2015) characterized machine learning as the "automation of automation"—a second-order automation that generates, rather than merely executes, algorithms. Vibe-Creation marks a further, qualitatively distinct stage: the automation of tacit knowledge, or what we term Vibe-Automation. In this process, the human's vague pre-reflective orientation meets the model's geometric sensitivity to directional pressure. The human provides the intentional impulse; the AI provides the geometric resolution; the Third Entity enacts the traversal of semantic space. The outcome is a configuration of meaning that neither participant could have specified in advance, because the specification is itself the product of the traversal.

This process bears a structural analogy to Polanyi's account of tacit knowledge. Just as tacit knowledge is the pre-reflective background that makes explicit specification possible for the individual human knower, the vibe is the pre-reflective background that makes the creative traversal of semantic space possible for the Third Entity. Vibe-Creation is, in this precise sense, the tacit dimension of human-machine cognition.

### 4.2 Three Historical Paradigms: The Michelangelo Example

The conceptual novelty of Vibe-Creation can be illuminated by tracing the evolution of creative agency through a classical artistic example. Consider the creation of Michelangelo's David, and analyze this creation across three historical paradigms.

In the first paradigm—Unified Creativity—the conceptual vision and the physical execution are fused within a single subject. Michelangelo stands before the marble block; the heroic form he perceives within it must be released through years of iterative manual labor, each chisel strike a micro-decision within a continuous negotiation between vision and material resistance. The scale of the creative act is bounded by the cognitive and physical bandwidth of a single embodied agent. The syntactic friction here is maximal: every insight must be translated into muscular action.

In the second paradigm—Collaborative Creativity—labor is divided between a visionary and an executor. Michelangelo retains the conceptual function while delegating physical execution to an ideal robotic sculptor. He issues instructions: "Carve a tensed youth with a sling on his shoulder, in contrapposto." The problem of physical friction is eliminated, but the paradigm remains epistemologically continuous with the first. The human thinks in the logic of control, issuing sequential symbolic commands to an apprentice who lacks independent judgment. The machine is faster and more precise, but the cognitive structure of the interaction is unchanged: a subject imposes a specification upon a compliant object.

In the third paradigm—Vibe-Creation—the structure of the interaction is transformed at the level of cognition itself. Michelangelo no longer issues instructions to an executor. He functions as a semantic architect navigating a fluid, high-dimensional geometric medium. Rather than specifying an outcome symbolically, he projects a constellation of semantic vectors—heroic anticipation, divine proportion, the synthesis of classical contrapposto and Renaissance humanism, the particular quality of arrested motion—and the form of David emerges as the resolution of the latent space under the pressure of this semantic gravity.

If the result is imperfect, the corrective gesture is not a micro-intervention (adjusting the angle of the chisel) but a macro-level reconfiguration of the vibe—adding, for instance, a vector of vulnerability—and the geometry of the figure reconstitutes accordingly. The creator does not construct an object; she curates the geometry of meaning.

This example makes visible the qualitative discontinuity between the three paradigms. In the third paradigm, the creator's agency is neither eliminated nor merely extended; it is transformed in kind. The relevant competence is no longer the ability to specify an outcome explicitly but the ability to orient a high-dimensional geometric process through pre-reflective attunement—to exercise what we propose to call navigational intelligence.

### 5. Asymmetric Emergence and the Question of Agency

The emergence of the Third Entity raises, with renewed urgency, the question of cognitive agency. If the outcomes of Vibe-Creation belong neither exclusively to the human participant nor exclusively to the AI system—if they are, in the strict sense, properties of the emergent coupled system—then the conventional locus of responsibility, authorship, and creative ownership is displaced. This question is not merely philosophical; it has immediate practical implications for attribution, accountability, and intellectual property in contexts where the Third Entity's outputs are consequential.

Two inadequate responses present themselves. The first is to locate agency entirely in the human, treating the AI as a sophisticated but ultimately instrumental executor of human intentions. This position preserves the familiar structure of accountability but does so at the cost of descriptive accuracy: it denies the constitutive role of the model's geometric navigation in determining what the human's intention actually becomes in the course of the interaction. The human who employs a generative AI system is not cognitively identical to the human who reasons without one; the encounter transforms the human's thinking, not merely its output.

The second inadequate response is to distribute agency equally between the two participants, treating the Third Entity as a partnership of peers. This position acknowledges the transformative character of the encounter but eliminates its fundamental asymmetry. The human brings to the interaction something that the AI system lacks absolutely: intentional orientation, ethical commitments, and existential stakes. The AI system has no investment in the outcome; it cannot be held responsible for consequences; it does not experience the encounter as a cognitive event. The asymmetry of existential involvement is not incidental to the structure of the Third Entity; it is constitutive of it.

The appropriate concept, we propose, is asymmetric emergence. The agency of the Third Entity is genuinely novel and irreducible to the prior capacities of either component; it cannot be fully attributed to either the human or the AI alone. But this emergent agency is not symmetrically distributed. It is anchored in human intentional responsibility: the human is the participant for whom the outcomes matter, who initiated the traversal through a desire for meaning, and whose evaluative judgment will determine the practical consequences of what was produced.

The human is not the sole author in any mechanically additive sense, but she is the author who matters—the one whose epistemic and ethical accountability is not discharged by pointing to the emergent structure of the interaction.

This notion of asymmetric emergence has implications beyond the individual act of co-creation. It suggests that the institutional structures we have developed to assign responsibility for intellectual work—authorship conventions, peer review, educational assessment, intellectual property regimes—require principled revision rather than ad hoc accommodation. The Third Entity is not a marginal or transitional phenomenon; it is, we argue, the characteristic form of advanced intellectual work in the age of generative AI.

## 6. Educational Implications: Toward a Pedagogy of Navigational Intelligence

The theoretical framework developed in the preceding sections—the Third Entity, Vibe-Creation, and asymmetric emergence—carries consequences that extend well beyond epistemology and the philosophy of mind. If the characteristic form of advanced cognitive work is now the navigational traversal of high-dimensional semantic space rather than the execution of explicit symbolic algorithms, then the institutions responsible for cultivating intellectual competence face a structural challenge that cannot be resolved by incremental curricular adjustment.

The classical research university was organized around the paradigm of overcoming syntactic friction. Engineering and humanities faculties alike trained students to be rigorous symbolic executors—to construct and manipulate well-formed symbol strings with precision, whether those strings took the form of code, mathematical proof, legal argument, or literary analysis. This training was pedagogically appropriate when the bottleneck of intellectual production was the human's capacity for explicit symbolic manipulation. That bottleneck has been substantially removed by generative AI systems capable of producing syntactically competent symbolic output at scale.

This shift from execution to navigation forces us to re-evaluate the foundational premises of educational technology. The implications extend to the philosophy of constructionism itself. Papert (1980) argued that the computer functions as an "object to think with"—a medium in which ideas can be externalized, tested, and refined through the act of construction. This insight retains its validity but requires extension: generative AI has evolved into an "agent to think with," a medium whose navigational properties actively co-constitute the ideas that emerge from the interaction, rather than merely providing a passive substrate for their externalization.

If this analysis is correct, then the primary educational challenge is no longer the cultivation of syntactic competence but the cultivation of navigational intelligence: the capacity to orient pre-reflective attunement, to exercise judgment about the directionality of semantic exploration, and to evaluate the outcomes of emergent co-creation from a position of epistemic and ethical accountability.

We propose the term Vibe-Engineer to designate the professional figure that educational institutions should now aim to produce—not a narrow specialist in a particular symbolic domain, but a practitioner capable of exercising macro-level conceptual direction over the Third Entity's navigational capacities.

The development of navigational intelligence requires, in turn, a reconceptualization of what we mean by intellectual assessment. Traditional assessment instruments—examinations, essays, code submissions—were designed to measure the quality of symbolic execution: the student's ability to produce correct, coherent, or original symbol strings under controlled conditions. These instruments are poorly suited to measuring navigational intelligence, which manifests not in the quality of individual symbolic outputs but in the quality of the directional impulses through which a student orients a high-dimensional co-creative process.

We suggest that the relevant competence—what might be termed Digital Intelligence (DQ)—involves at minimum the following capacities: the ability to formulate navigational vectors in high-dimensional semantic space with sufficient precision to produce epistemically interesting outputs; the ability to evaluate emergent outputs against the background of the vibe that generated them, distinguishing productive from unproductive resolutions; and the ability to assume the asymmetric ethical responsibility for outcomes that the structure of the Third Entity places, irreducibly, upon the human participant.

## 7. Directions for Future Research

The theoretical framework proposed here opens several urgent avenues for future investigation:

- Psychological Specification: The concept of navigational intelligence demands detailed psychological mapping, grounding it in empirical research on human-computer interaction and cognitive load during Vibe-Creation.
- Operationalizing DQ: The notion of Digital Intelligence requires translation into validated pedagogical frameworks and assessment instruments capable of measuring navigational competence rather than syntactic execution.
- Embodied Cognition: The postphenomenological analysis of the Third Entity invites sustained engagement with the tradition of embodied cognition (Varela et al., 1991; Clark & Chalmers, 1998), exploring how the coupled system's cognitive properties remain physically anchored in the human user.
- Ethical Frameworks: The concept of asymmetric emergence necessitates the development of new institutional and legal frameworks to address the displacement of traditional authorship and intellectual property in AI-mediated research.

## 8. Conclusion

This paper has argued that the encounter between human reasoning and generative AI produces a qualitatively distinct cognitive-epistemic formation—the Third Entity—that cannot be adequately theorized within the inherited frameworks of tool-use, augmentation, or collaborative partnership. Drawing on a multi-disciplinary theoretical architecture, we have characterized this formation as an emergent, transient structure produced by the transductive coupling of two ontologically incommensurable cognitive modes: the symbolic-intentional mode of the human and the geometric-indexical mode of the AI system.

We have introduced the concept of Vibe-Creation to designate the cognitive mode through which the Third Entity operates, argued that this mode constitutes the automation of tacit knowledge, and proposed the notion of asymmetric emergence to characterize the agency of the resulting formation. We have further argued that this theoretical framework carries substantial implications for educational institutions, which must now address the challenge of cultivating navigational intelligence—the capacity to orient high-dimensional co-creative processes from a position of genuine epistemic and ethical accountability.

What can be stated with some confidence at this stage is the following: the categories we have inherited for describing the relationship between human intelligence and computational technology—tool, collaborator, assistant, oracle—are inadequate to the phenomenon that generative AI has produced. The Third Entity does not yet occupy a legitimate place in our epistemology, our ethics, or our educational institutions. The present article is an attempt to begin constructing one.

## REFERENCES

Bengio, Y., Courville, A., & Vincent, P. (2013). Representation learning: A review and new perspectives. *IEEE Transactions on Pattern Analysis and Machine Intelligence, 35*(8), 1798–1828.

Clark, A., & Chalmers, D. (1998). The extended mind. *Analysis, 58*(1), 7–19.

Davis, N., Hsiao, C. P., Singh, K. Y., Li, L., & Magerko, B. (2013). An enactive model of creativity for computational collaboration and co-creation. In T. Veale, K. Feyaerts, & C. Forceville (Eds.), *Creativity and the agile mind* (pp. 259–281). De Gruyter Mouton.

Domingos, P. (2015). *The master algorithm: How the quest for the ultimate learning machine will remake our world*. Basic Books.

Heidegger, M. (1962). *Being and time* (J. Macquarrie & E. Robinson, Trans.). Harper & Row. (Original work published 1927)

Hutchins, E. (1995). *Cognition in the wild*. MIT Press.

Ihde, D. (1990). *Technology and the lifeworld: From garden to earth*. Indiana University Press.

Kantosalo, A., & Toivonen, H. (2016). Modes and styles of collaboration in co-creative systems. In *Proceedings of the AAAI Workshop on Intelligent Narrative Technologies*.

Levin, I. (2026). Epistemology of generative AI: The geometry of knowing. *arXiv preprint arXiv:2602.17116*.

Mikolov, T., Sutskever, I., Chen, K., Corrado, G. S., & Dean, J. (2013). Distributed representations of words and phrases and their compositionality. *Advances in Neural Information Processing Systems, 26*.

Morin, E. (1977). *La méthode, tome 1: La nature de la nature*. Éditions du Seuil.

Papert, S. (1980). *Mindstorms: Children, computers, and powerful ideas*. Basic Books.

Peirce, C. S. (1931–1958). *Collected papers of Charles Sanders Peirce* (Vols. 1–8, C. Hartshorne, P. Weiss, & A. Burks, Eds.). Harvard University Press.

Polanyi, M. (1966). *The tacit dimension*. Doubleday & Company.

Simondon, G. (1958a). *L'individuation à la lumière des notions de forme et d'information*. Presses Universitaires de France.

Simondon, G. (1958b). *Du mode d'existence des objets techniques*. Éditions Aubier-Montaigne.

Varela, F. J., Thompson, E., & Rosch, E. (1991). *The embodied mind: Cognitive science and human experience*. MIT Press.

Wellner, G., & Levin, I. (2024). Ihde meets Papert: Combining postphenomenology and constructionism for a future agenda of philosophy of education in the era of digital technologies. *Learning, Media and Technology*, 1–14.

Wellner, G., & Levin, I. (2025). A postphenomenological-constructionist assessment of AI in education toward a multidimensional democratic education. In M. Bohlmann & P. Breil (Eds.), *Postphenomenology and technologies within educational settings* (pp. 225–241). Lexington Books.